\documentclass[a4paper]{IEEEtran}
\usepackage{hyperref}

\usepackage[usenames, dvipsnames]{color}
\usepackage{amstext,amsfonts,amssymb,amsthm,amsmath,mathrsfs}
\usepackage{amstext,amsfonts,amssymb,amsthm,amsmath,mathrsfs}
\usepackage[english]{babel}

\usepackage{epstopdf}
\usepackage{graphicx}
\usepackage{float}

\usepackage{multicol}

\makeglossary
\newlength{\nmlw}

\newcommand{\comment}[1]{}

\renewcommand{\eqref}[1]{Eq.~(\ref{#1})}

\newcommand{\figref}[1]{Fig.~\ref{#1}}

\newcommand*{\rom}[1]{\expandafter\@slowromancap\romannumeral #1@}

\title{Deep Learning based Estimation of Weaving Target Maneuvers}

\author{Vitaly Shalumov\IEEEauthorrefmark{1}\thanks{\IEEEauthorrefmark{1}Rafael, vitalysh@rafael.co.il}
	\ and
	Itzik Klein\IEEEauthorrefmark{2}\thanks{\IEEEauthorrefmark{2}Dr., Research fellow, Rafael, itzikkl@rafael.co.il}\\
	{\normalsize\itshape
		Rafael, P.O.Box 2250, Haifa 3102102, Israel}\\
}

\begin{document}
\maketitle
\begin{abstract}
In target tracking, the estimation of an unknown weaving target frequency is crucial for improving the miss distance. The estimation process is commonly carried out in a Kalman framework. The objective of this paper is to examine the potential of using neural networks in target tracking applications.  To that end, we propose estimating the weaving frequency using deep neural networks, instead of classical Kalman framework based estimation. Particularly, we focus on the case where a set of possible constant target frequencies is known. Several neural network architectures, requiring low computational resources were designed to estimate the unknown frequency out of the known set of frequencies. The proposed approach performance is compared with the multiple model adaptive estimation algorithm. Simulation results show that in the examined scenarios, deep neural network outperforms multiple model adaptive estimation in terms of accuracy and the amount of required measurements to convergence.
\end{abstract}

\section{Introduction} \label{sec:Introduction}
The problem of tracking maneuvering targets has been studied extensively in the literature, including in terms of performance \cite{bar2004estimation,bar2011tracking} and observability \cite{klein2014observability}. Most target maneuver models are turn or constant motion models and are  established relying on the target expected kinematics. When considering the problem of estimating the target position while maneuvering in a barrel roll or coordinated turn, the unknown frequency of the target maneuver needs to be estimated. In such situations, the target dynamics is commonly simplified and modelled as a sine wave with a constant frequency.  This frequency can be known or unknown from a given target frequencies set.
If the target frequencies set is unknown, approaches such as Least Squares (LS), linear  Kalman Filter (KF), Extended Kalman Filter (EKF) and State Dependent Differential-Difference Riccati Equation can be applied to estimate the frequency \cite{zarchan2005progress,rusnak2012new}. On the other hand, if the set of possible target frequencies is known, the problem can be formulated as a estimation problem of the true  target frequency out from a known set of possible frequencies.
Several approaches exist in the literature for solving the latter problem. Among them are the  Multiple Model Adaptive Estimator (MMAE) \cite{magill1965optimal,sims1969recursive} and Interacting Multiple Model (IMM) \cite{blom1984efficient,blom1988interacting}; both of them are derived in the KF framework. The MMAE algorithm assumes that the estimated frequency belongs to a known closed set of frequencies. There, each frequency is tuned to a different KF/EKF model. The MMAE uses this bank of Kalman filters to estimate the target frequency by updating the probability of each model being the correct one using a likelihood function and the Bayes rule. For example, \cite{marks2006multiple} and \cite{zarchan2006using} implemented MMAE filter for improving guidance performance against weaving targets.
Another approach implemented in the literature is IMM, which uses a switching multiple model approach. In contrast to MMAE, this approach enables several different models to be the true model at different times. The IMM algorithm is more complex than the MMAE algorithm, yet it can be shown that the MMAE is a subset of the IMM algorithm \cite{zarchan2005progress}.

In parallel to the progress in target tracking theory, major progress was made in the field of machine learning and in particular deep learning (DL). 
These methods are actually application of artificial neural network (ANN) \cite{mcculloch1943logical}, with a deep architecture, i.e. with several hidden layers. The common DL architectures are fully connected neural network, known as the convolutional neural networks (CNN) and the Recurrent Neural Network (RNN) used for time-sequences. In the scope of time series classification (i.e., using time varying signals), CNNs were used to classify audio signals \cite{lee2009unsupervised}. Ref.\cite{jiang2015human} converted signals produced by wearable sensors into images, and used a CNN for classification. In  Ref.\cite{gao2016deep}, a Long Short-Term Memory (LSTM) network, which is an extension of RNNs, was compared to a CNN, in terms of classification performance of Visual and Haptic Data in a robotics setting.
Ref.\cite{wang2015encoding} presented two approaches for transformation of the time series into an image: the Gramian Angular Field (GAF) and the Markov Transition Field (MTF). The authors suggested combining both methods and using them as "colors" of the images.

In this paper, we propose to merge the fields of target tracking and machine learning by deriving a deep artificial neural network (DNN) to estimate the unknown constant target frequency out of a prior set of known frequencies. To that end, we design an appropriate network and compare our proposed approach with the classical MMAE estimation algorithm. Simulation results show that our proposed DNN approach outperforms the MMAE algorithm in the examined scenarios in both accuracy and convergence time.

The remainder of the paper is organized as follows: Section \ref{sec:Popular Frequency Identification Estimators} presents the formulation of the problem and the MMAE approach. Section \ref{sec:Frequency Identification} presents our DNN estimator while in Section \ref{sec:Simulations} a comparison is made between the DL and MMAE algorithms. Finally, Section \ref{sec:Conclusions} shows the main conclusions of the paper.

\section{Multiple Model Adaptive Estimator}
\label{sec:Popular Frequency Identification Estimators}
Let the target trajectory be modeled as a one dimensional sine wave. The noisy measurement of such a trajectory is given by:
\begin{equation}\label{eq:1}
z(t) = A\sin (\omega t + \phi ) + v,\;\;\;t \in \left[ {0,T} \right]
\end{equation}
where $A$ is the amplitude, $\omega$ is a constant frequency,
  $\phi$ is the phase, $T$ is the time at the end of the measurement sequence, $v$ is the measurement noise modeled as zero mean white Gaussian noise:
\begin{equation}\label{eq:20}
v \sim N\left( {0,{\sigma ^2}} \right)
\end{equation}
and $\sigma^2$ is the measurement variance.

We consider the case in which the frequency of the target sine maneuver is unknown, but constant throughout the measured sequence and the identification of the frequency is required only at the end of the of the measured time sequence i.e. at $t=T$. It is further assumed that the frequency of the target, $\omega$, is in a known set $\Omega $, such that $\omega  \in \Omega $. A common approach to address such a problem is by using a bank of linear two state Kalman filters. Such a filter is presented in the following subsection, adapted from Ref.\cite{zarchan2005progress}.


%

\subsection{Linear Two State Sine Wave Kalman Filter}
\label{subsubsec:LKF}
Let the position of the target be described by:
\begin{equation}\label{eq:2}
x = A\sin (\omega t+\phi)
\end{equation}
After taking a double derivative, we obtain the following equation:
\begin{equation}\label{eq:3}
\ddot x =  - {\omega ^2}x
\end{equation}
or in a state-space form:
\begin{equation}\label{eq:4}
\frac{d}{{dt}}\left[ {\begin{array}{*{20}{c}}
x\\
{\dot x}
\end{array}} \right] = \underbrace {\left[ {\begin{array}{*{20}{c}}
0&1\\
{ - {\omega ^2}}&0
\end{array}} \right]}_{\bf{F}}\left[ {\begin{array}{*{20}{c}}
x\\
{\dot x}
\end{array}} \right]
\end{equation}
The transition matrix of a linear time invariant system for a continuous case is given by:
\begin{equation}\label{eq:5}
{\bf{\Phi }}(t) = {\mathcal{L}^{ - 1}}\left[ {{{\left( {s{\bf{I}} - {\bf{F}}} \right)}^{ - 1}}} \right] = \left[ {\begin{array}{*{20}{c}}
{\cos \omega t}&{\frac{{\sin \omega t}}{\omega }}\\
{ - \omega \sin \omega t}&{\cos \omega t}
\end{array}} \right]
\end{equation}
or for discrete case as:
\begin{equation}\label{eq:6}
{{\bf{\Phi }}_k} = \left[ {\begin{array}{*{20}{c}}
{\cos \omega {T_s}}&{\frac{{\sin \omega {T_s}}}{\omega }}\\
{ - \omega \sin \omega {T_s}}&{\cos \omega {T_s}}
\end{array}} \right]
\end{equation}
The discrete process noise covariance is given by:
\begin{equation}\label{eq:7}
{{\bf{Q}}_k} = \int\limits_0^{{T_s}} {{\bf{\Phi }}(\tau )} {\bf{Q}}{{\bf{\Phi }}^T}(\tau )dt,\;{\bf{Q}} = \left[ {\begin{array}{*{20}{c}}
0&0\\
0&{{\Phi _s}}
\end{array}} \right]
\end{equation}
where $\bf{Q}$ is the corresponding continuous process noise with a continuous process noise spectral density $\Phi _s$.\\
The linear Kalman filter is given by:
\begin{equation}\label{eq:8}
{{{\bf{\hat x}}}_k} = {{\bf{\Phi }}_k}{{{\bf{\hat x}}}_{k - 1}} + {{\bf{K}}_k}\left[ {{z_k} - {\bf{H}}{{\bf{\Phi }}_k}{{{\bf{\hat x}}}_{k - 1}}} \right]
\end{equation}

\begin{subequations}\label{eq:9}
\begin{alignat}{3}
{{\bf{M}}_k} &= {{\bf{\Phi }}_k}{{\bf{P}}_{k-1}}{\bf{\Phi }}_{_k}^T + {{\bf{Q}}_k}\\
{{\bf{K}}_k} &= {{\bf{M}}_k}{{\bf{H}}^T}{({\bf{H}}{{\bf{M}}_k}{{\bf{H}}^T} + {{\bf{R}}_k})^{ - 1}}\\
{{\bf{P}}_k} &= ({\bf{I}} - {{\bf{K}}_k}{\bf{H}}){{\bf{M}}_k}
\end{alignat}
\end{subequations}
where ${{\bf{M}}_k}$ is the covariance matrix representing errors in the state estimates before an update, ${{\bf{K}}_k}$ is the Kalman gain, ${{\bf{P}}_k}$ is a covariance matrix representing errors in the state estimates, $R$ is the measurement noise covariance, taken according to \eqref{eq:20}, and the measurement $z(t)$ is given in \eqref{eq:1} with the measurement matrix taken as:
\begin{equation}\label{eq:10}
{\bf{H}} = \left[ {\begin{array}{*{20}{c}}
1&0
\end{array}} \right]
\end{equation}

\subsection{Sine Wave MMAE}
\label{subsubsec:Sine Wave MMAE}

The linear two state Kalman filter assumes an unknown, but constant target frequency. The bank of filters is constructed by assigning to each two state linear Kalman filter a different frequency out of the set $\Omega$. Each model receives an initial probability being the correct model and the MMAE updates those probabilities.

Let $N$ be the number of models in the bank of filters. The i-th filter residual (${{\mathop{\rm Res}\nolimits} _k}(i)$) and covariance ($\sigma _{{{{\mathop{\rm Res}\nolimits} }_k}}^2(i)$) at the k-th instant are defined as follows:
\begin{subequations}\label{eq:15}
\begin{alignat}{2}
{{\mathop{\rm Res}\nolimits} _k}(i) &= {{{z}}_k} - {\bf{H}}{{\bf{\Phi }}_k}(i){{{\bf{\hat x}}}_{k - 1}}(i)\\
\sigma _{{{{\mathop{\rm Res}\nolimits} }_k}}^2(i) &= {C_k}(i) = {\bf{H}}{{\bf{M}}_k}(i){{\bf{H}}^T} + {{{R}}_k}
\end{alignat}
\end{subequations}
The likelihood function for the i-th filter at the k-th instant is given as follows:
\begin{equation}\label{eq:16}
{f_k}(i) = \frac{1}{{\sqrt {2\pi {C_k}(i)} }}\exp \left[ { - 0.5{\mathop{\rm Res}\nolimits} _k^2(i)/{C_k}(i)} \right]
\end{equation}
The probability that the i-th filter is the correct one at the k-th instant is calculated by:
\begin{equation}\label{eq:17}
{p_k}(i) = \frac{{{f_k}(i){p_{k - 1}}(i)}}{{\sum\limits_{i = 1}^r {{f_k}(i){p_{k - 1}}(i)} }}
\end{equation}
Usually, by assuming that each filter has the same initial probability to be the correct one, we use
\begin{equation}\label{eq:18}
{p_0}(i) = \frac{1}{N}
\end{equation}
The MMAE estimated state, ${{{\bf{\hat x}}}_{MMAE,k}}$, is calculated as a weighted sum of the estimate state of each filter state (\eqref{eq:8}) in which the weights are the corresponding probability:
\begin{equation}\label{eq:19}
{{{\bf{\hat x}}}_{MMAE,k}} = \sum\limits_{i = 1}^r {{p_k}(i){{{\bf{\hat x}}}_k}(i)}
\end{equation}
The estimated  frequency of the target is:
\begin{equation}\label{eq:21}
{\omega _k} = \sum\limits_{i = 1}^r {{p_k}(i)\omega (i)}
\end{equation}
That is, the estimated frequency of the target is the resulting sum of the filter probability being the correct one multiplied by its assigned target frequency.

\section{Frequency Identification using Deep Neural Networks}
\label{sec:Frequency Identification}

\subsection{Background}
\label{subsec:Frequency Identification Background}

An Artificial Neural Network (ANN) is a biologically inspired simulation framework, that can perform task such as classification, pattern recognition, neutral language processing and more.
An ANN is a network of computing units called "neurons", linked by connections called "synapses". Each layer of the network, consisted of several neurons, is connected to the next layer, with the connection having a weight that corresponds to the strength of the connection \cite{gamboa17}.
Each neuron in a layer typically receives the sum of all connections from the previous layer, passed through an activation function.
Among the most popular activation functions are the logistic sigmoid
\begin{equation}
 \sigma (x) = \frac{1}{{1 + {e^{ - x}}}},
\end{equation}
the hyperbolic tangent
\begin{equation}
\tanh (x) = \frac{2}{{1 + {e^{ - 2x}}}} - 1
\end{equation}
and the rectified linear unit
\begin{equation}
R(x) = \max (0,x)
\end{equation}
known as ReLU. For example, the value of the third neuron in the second layer, $x_{3,L2}$, which is connected to two neurons from the first layer, $x_{1,L1}$ and  $x_{2,L1}$, passed through the ReLU activation function is calculated as follows:
\begin{equation}\label{eq:22}
{x_{3,L2}} = \max \left( {0,{w_0} + {w_1}{x_{1,L1}} + {w_2}{x_{2,L1}}} \right)
\end{equation}
where $w$-s are the synapses' weights.

\begin{figure}[!ht]
	\centering
	\includegraphics[width=6cm]{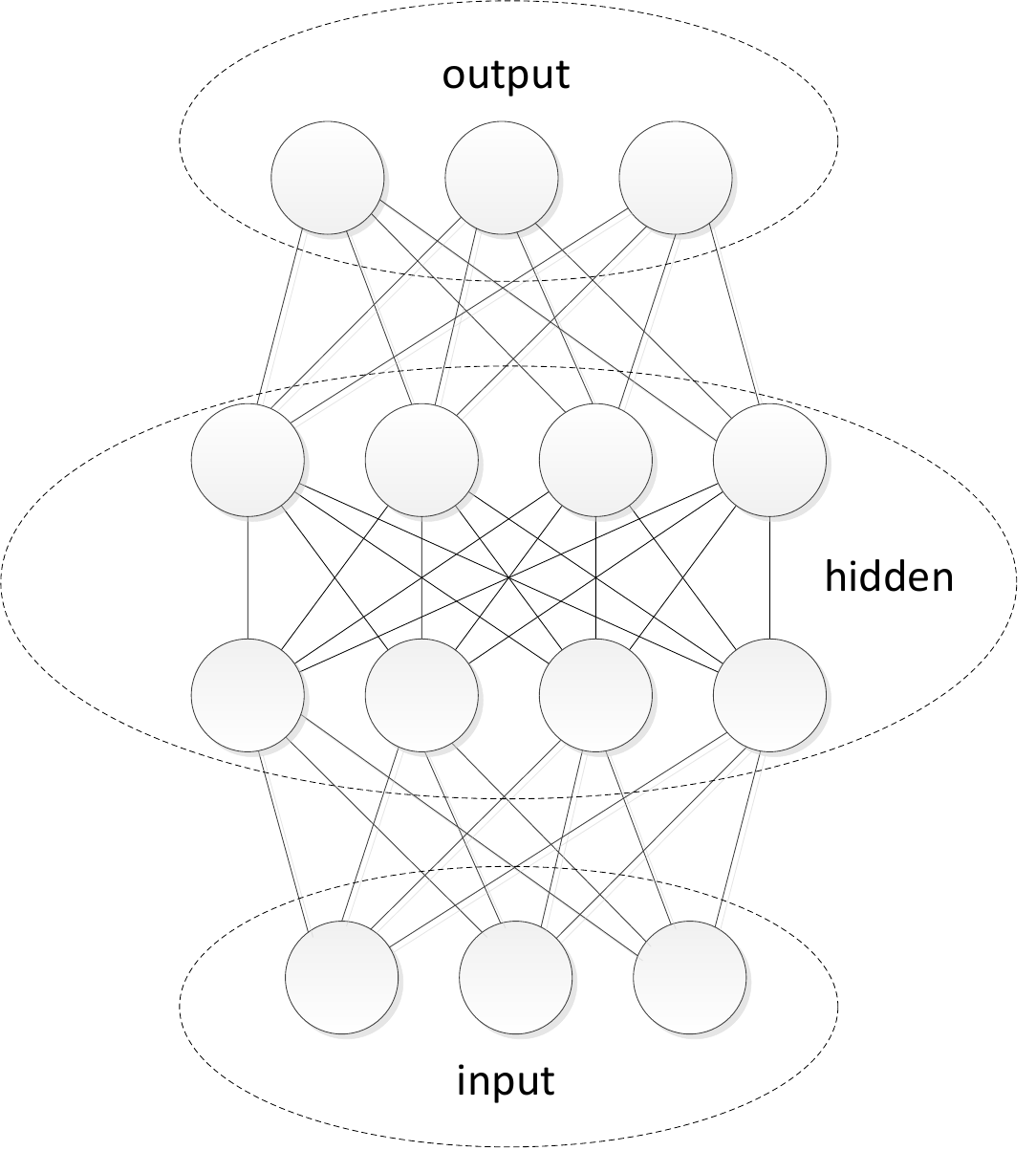} \\
	\caption{An example for artificial neural network, presenting an input, output and hidden layers.}
	\label{fig:NN}
\end{figure}

The input to the network consists of an input layer and output layer where every other layer in between is defined as \emph{hidden layers}. A DNN is actually an ANN, with a deep architecture, i.e. several hidden layers. The objective of the network is to calculate the weights that will, given the input, predict the desired output. For example in this paper, when given the target measurements as input to the network, the goal is to output the target frequency.
\figref{fig:NN} presents a typical ANN with an input and output layers of three neurons, two hidden layer, each with four neurons, where all the layers are fully connected (adapted from \cite{gamboa17}).

At each iteration, the network updates the weights as to minimize a given loss function, which reflects how far is the predicted output from the desired output. The most popular optimization method in neural networks is the gradient descent.
Gradient descent minimizes an objective function $J(\theta )$ ($\theta$ are the weights), by updating the parameters in the opposite direction of the gradient of the objective function. Three gradient descent variants \cite{ruder2016overview}, which differ in the amount of data used to compute the gradient are the Batch gradient descent:
 \begin{equation}\label{eq:12}
 \theta  = \theta  - \eta {\nabla _\theta }J(\theta )
 \end{equation}
 Stochastic gradient descent:
 \begin{equation}\label{eq:13}
 \theta  = \theta  - \eta {\nabla _\theta }J(\theta ;{x^{(i)}};{y^{(i)}})
 \end{equation}
 and the Mini-batch gradient descent:
 \begin{equation}\label{eq:14}
 \theta  = \theta  - \eta {\nabla _\theta }J(\theta ;{x^{(i:i + n)}};{y^{(i:i + n)}})
 \end{equation}
where $x$ and $y$ are the training set and its labels respectively, $i$ is the index of the particular training example and its label and $n$ is the size of the mini-batch.
All of those approaches use a constant learning rate for the optimization. In some situations, there is a need to use an adaptive learning rate value to obtain good results with less computation time. Two common methods that use an adaptive learning rate as an extension to gradient descent are Adam\cite{kingma2014adam} and Adagrad\cite{duchi2011adaptive}.

\begin{figure}[!ht]
  \centering
  \includegraphics[width=8cm]{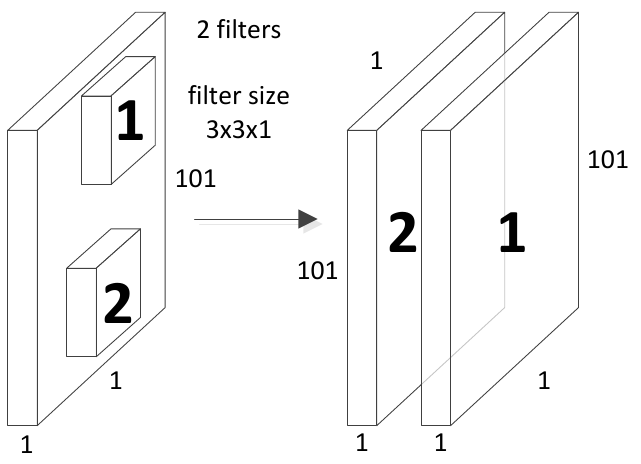} \\
  \caption{An example of a convolutional layer, showing two filters, as applied in the paper.}
  \label{fig:CNN}
\end{figure}

In addition, convolutional neural networks, commonly used for image classification, use filters to create convolutional layers. In CNNs, the connections between neurons are limited by design in a way that allows the use of smaller number of tunable parameters i.e. weights. Each convolution between a local area of the image and the filter extracts important features from the image. The convolution operation is essentially a dot product between the filter weights and the local area of the image.
It is sometimes convenient to keep the dimensions of the filter application result equal to the dimensions of the layer on which the convolution is performed. To this end, the layer undergoing the convolution is padded with zeros, in a process knows as "padding".
\figref{fig:CNN} presents a convolutional layer with two filters, each of size 3x3x1.

\subsection{Chosen DNN Architecture}
\label{subsec:Chosen DNN Architecture}
For simplicity, we assume that all target measured data has the same time length $T$. This assumption enables to treat the time series measurements as a constant length vector as input to the neural network architectures. The goal of the network is to output the correct target frequency (true label, in machine learning terminology).

Four types of network architectures were designed in order to perform the target frequency classification task.  Two of those networks are fully connected as presented in Table \ref{DNN_Table} and the other two are CNNs as presented in Table \ref{CNN_Table}.

\begin{table}[!ht]
  \caption{CNN network parameters.}\label{CNN_Table}
  \centering{\resizebox{1\hsize}{!}{$
\begin{array}{*{20}{c}}
\underline{Name}&\underline{Filters}&\begin{array}{l}
\underline{KernelSize},\\
\underline{Stride}
\end{array}&\begin{array}{l}
\underline{Padding},\\
\underline{MaxPooling}
\end{array}&\underline{Activation}&\underline{Optimizer}\\
{CNN}&{4,8,12,16}&{3,1}&{Yes,No}&{{\mathop{\rm ReLU}\nolimits} }&{Adam}\\
{CNN2}&\begin{array}{l}
\\
4,8,12,16,\\
20,24,28
\end{array}&{3,1}&{Yes,No}&{{\mathop{\rm ReLU}\nolimits} }&{Adam}
\end{array}$}}
\end{table}

\begin{table}[!ht]
  \caption{Fully connected network parameters.}\label{DNN_Table}
  \centering{
$\begin{array}{*{20}{c}}
\underline{{Name}}&\underline{{Layers}}&\underline{{Activation}}&\underline{{Optimizer}}\\
{DNN}&{10,20,10}&{{\mathop{\rm ReLU}\nolimits} }&{Adagrad}\\
{DNN2}&{40,30,20,10}&{{\mathop{\rm ReLU}\nolimits} }&{Adagrad}
\end{array}$}
\end{table}

These network architectures were chosen due to their structure simplicity and low computational resources required for training (each network trained for less than a minute on a desktop computer without a GPU).

For the CNNs, we chose to apply pooling on the convolutional layers, thus reducing the size of the network and yet retain its important features. A common pooling method named MaxPooling was used, which takes the maximum value from a given size matrix, which is obtained by sliding a window on the convolutional layer.

In addition, we used the Adam and Adagrad gradient descent algorithms. Adam, which stands for adaptive moment estimation calculates ${m_t}$ which is the  exponentially decaying average of past gradients and ${v_t}$ which is the exponentially decaying average of past squared gradients:
\begin{subequations}\label{eq:23}
\begin{alignat}{2}
\begin{array}{l}
{m_t} = {\beta _1}{m_{t - 1}} + (1 - {\beta _1}){g_t}\\
{v_t} = {\beta _2}{v_{t - 1}} + (1 - {\beta _2})g_{_t}^2
\end{array}
\end{alignat}
\end{subequations}
which are the estimation of the first two moments of the gradients. $g_t$ is the vectorization of $g_{t,i}$ that is calculated as follows:
\begin{equation}\label{eq:26}
{g_{t,i}} = {\nabla _{{\theta _t}}}J({\theta _{t,i}})
\end{equation}

To counteract the biases of  ${m_t}$  and  ${v_t}$  (these moments are biased towards zero), the moments are corrected as follows:
\begin{subequations}\label{eq:24}
\begin{alignat}{2}
\begin{array}{l}
{{\hat m}_t} = \frac{{{m_t}}}{{1 - \beta _1^t}}\\
{{\hat v}_t} = \frac{{{v_t}}}{{1 - \beta _2^t}}
\end{array}
\end{alignat}
\end{subequations}

The parameters are then updated by the following rule:
\begin{equation}\label{eq:25}
{\theta _{t + 1}} = {\theta _t} - \frac{\eta }{{\sqrt {{{\hat v}_t} + \varepsilon } }}{{\hat m}_t}
\end{equation}
The suggested algorithm parameters values are: $\beta _1=0.9$, $\beta _2=0.999$ and $\epsilon=10^{-8}$:

Adagrad, which is suited for dealing with sparse data, due to the fact that it performs larger updates for infrequent parameters and smaller for the frequent ones, has the following update rule:
\begin{equation}\label{eq:30}
{\theta _{t + 1}} = {\theta _t} - \frac{\eta }{{\sqrt {{G_t} + \varepsilon } }} \odot {g_t}
\end{equation}
where $G_t$ is a diagonal matrix, where each diagonal element is the sum of the squares of the gradients w.r.t. $\theta _i$ up to time $t$, and $\odot$ is an element-wise-matrix-vector multiplication.
$\epsilon$ is of the order of $10^{-8}$ and  $\eta$ is commonly taken as $0.01$.

The data set to the neural networks estimator consisted of a sine wave with length of $T$ constructed  with a sampling time of $0.01 [s]$ using \eqref{eq:1} while assuming zero phase. For each frequency in the known set of frequencies, 3000 sine wave signals were created , i.e. for the case of three frequencies, the data set consisted of 9000 signals. The dataset was divided to training dataset and test dataset where the training data consisted of 8000 signals (about $89\%$ of the data) and the test data consisted of 1000 signals (about $11\%$ of the data).

%

\section{Analysis and Discussion}
\label{sec:Simulations}
A comparison of the performance between neural networks architectures and MMAE estimators, using a numerical simulation, for a set of known frequencies $\Omega $ consisting of three frequencies is made. Note, that due to the fact that we address a classification problem (classify the correct frequency), the terms estimation (used in target tracking terminology) and classification (used in machine learning terminology) will be used interchangeably throughout this section since they have the same meaning in this work.

All four neural networks architectures as described in Tables 1-2 obtained almost the same performance. Therefore, for clarity of the results we present in the following figures only the comparison between the MMAE estimator and the DNN architecture as described in Table 2, line 1 (a fully connected neural network with three hidden layers, having 10, 20, and 10 hidden units ("neurons") respectively, with a ReLU activation for each layer, with a softmax cross entropy loss function and with an Adagrad optimizer).

\begin{figure}[!ht]
  \centering
  \includegraphics[width=9cm]{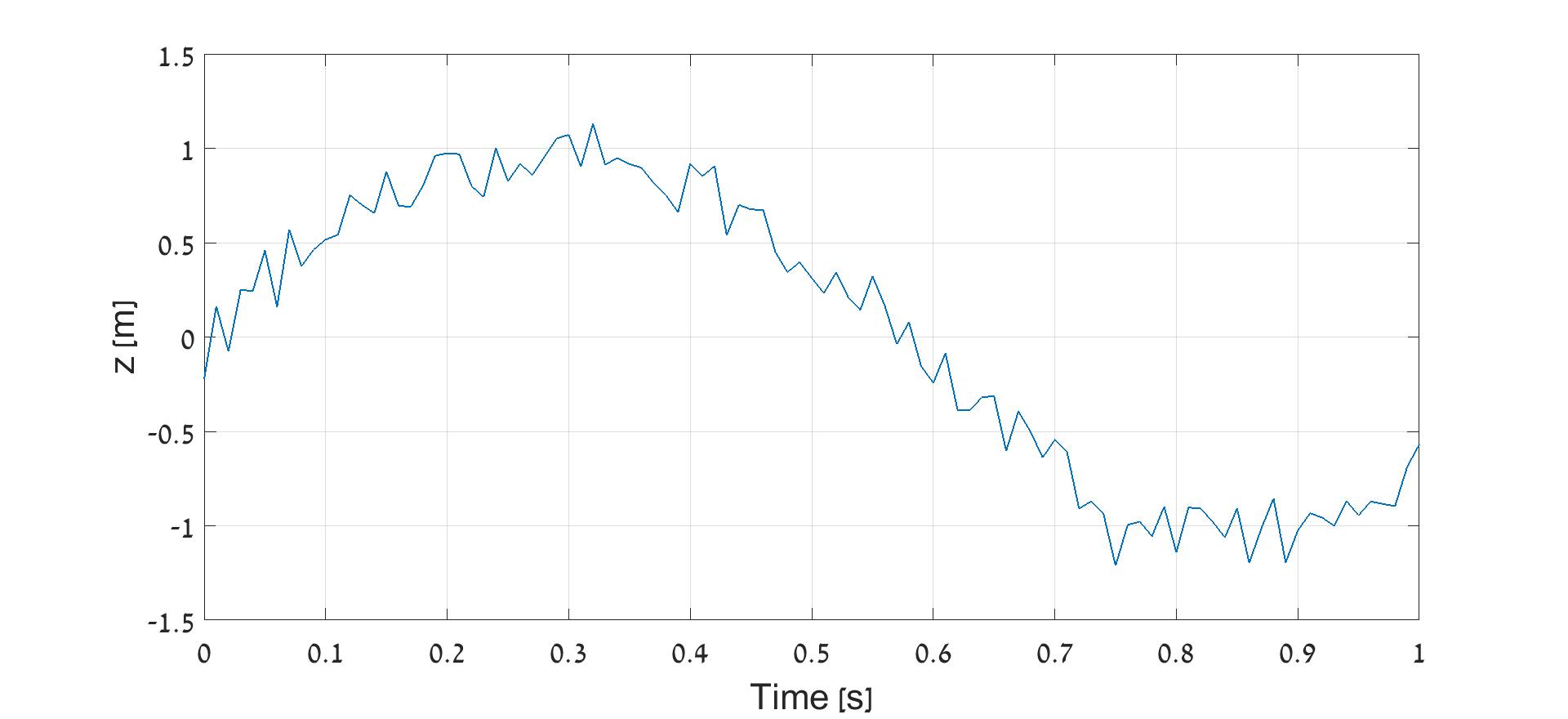} \\
  \caption{Sine wave example.}
  \label{fig:sin}
\end{figure}

The performance is evaluated for three sets of frequencies varying in their magnitudes and the difference between frequencies in the set:
\begin{subequations}\label{eq:11}
\begin{alignat}{3}
&{\Omega _1} = \left\{ {5,5.5,6} \right\}[Hz]\\
&{\Omega _2} = \left\{ {5,5.2,5.4} \right\}[Hz]\\
&{\Omega _3} = \left\{ {10,10.2,10.4} \right\}[Hz]
\end{alignat}
\end{subequations}
A sampling time of $0.01 [s]$ and zero phase ($\phi=0[rad]$) were used to create the dataset. For example, \figref{fig:sin} presents a sine wave with $T=1 [s]$, a constant frequency, and noise standard deviation of $\sigma =0.1 [m]$.

In order to compare both estimators, the accuracy parameter was chosen as a measure of performance.
The accuracy is calculated by the number of test signals classified correctly, divided by the total number of test samples.
The MMAE classification was performed by running an MMAE estimator on each signal in the test set, and the Kalman filter with the highest probability at $t=T$ was taken as the estimator's classification.

\begin{figure}[!ht]
  \centering
  \includegraphics[width=9cm]{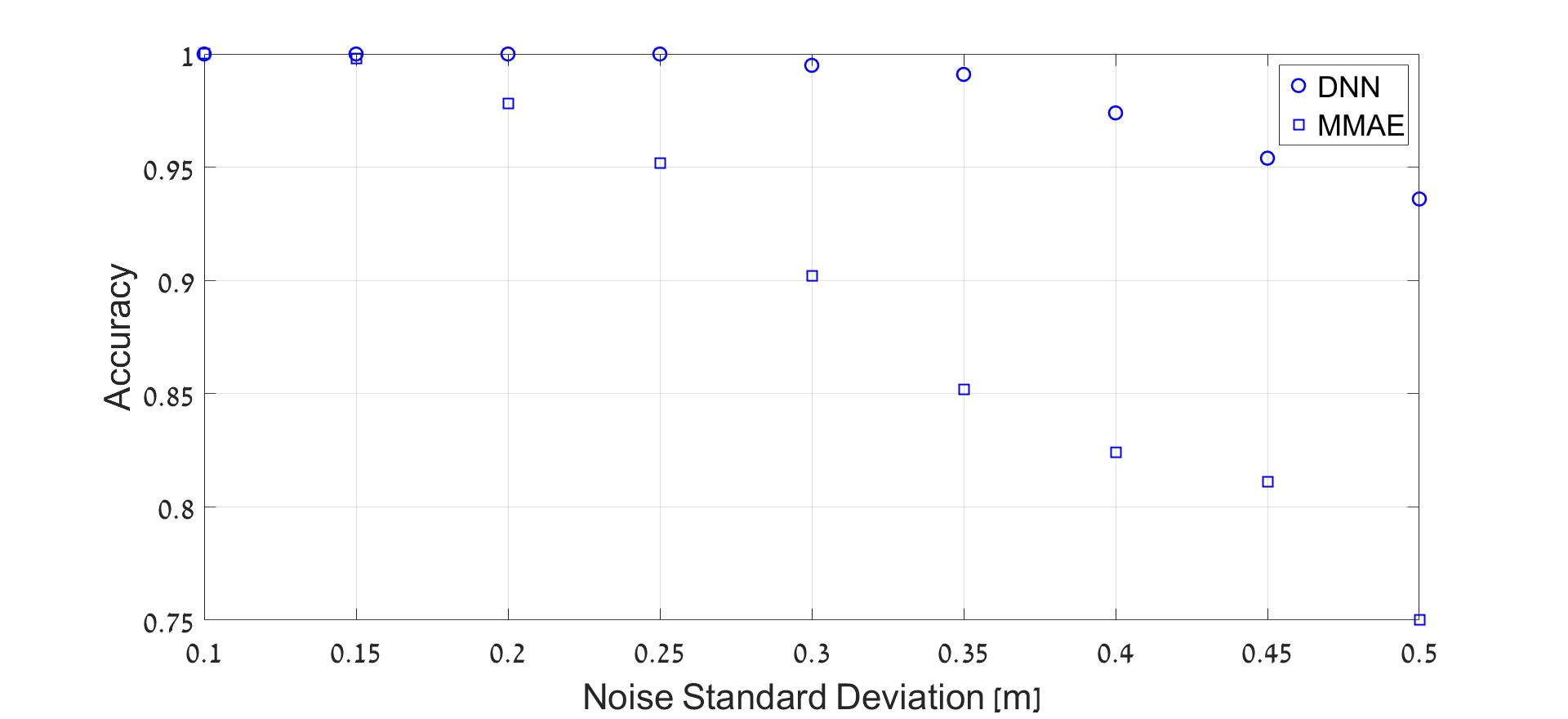} \\
  \caption{Accuracy vs noise standard deviation for \newline $\omega  \in \left\{ {5,5.5,6} \right\}\;{\mkern 1mu} [Hz]\;(\omega  \in {\Omega _1})$.}
  \label{fig:Acc_DNN_MMAE_5}
\end{figure}

\begin{figure}[!ht]
  \centering
  \includegraphics[width=9cm]{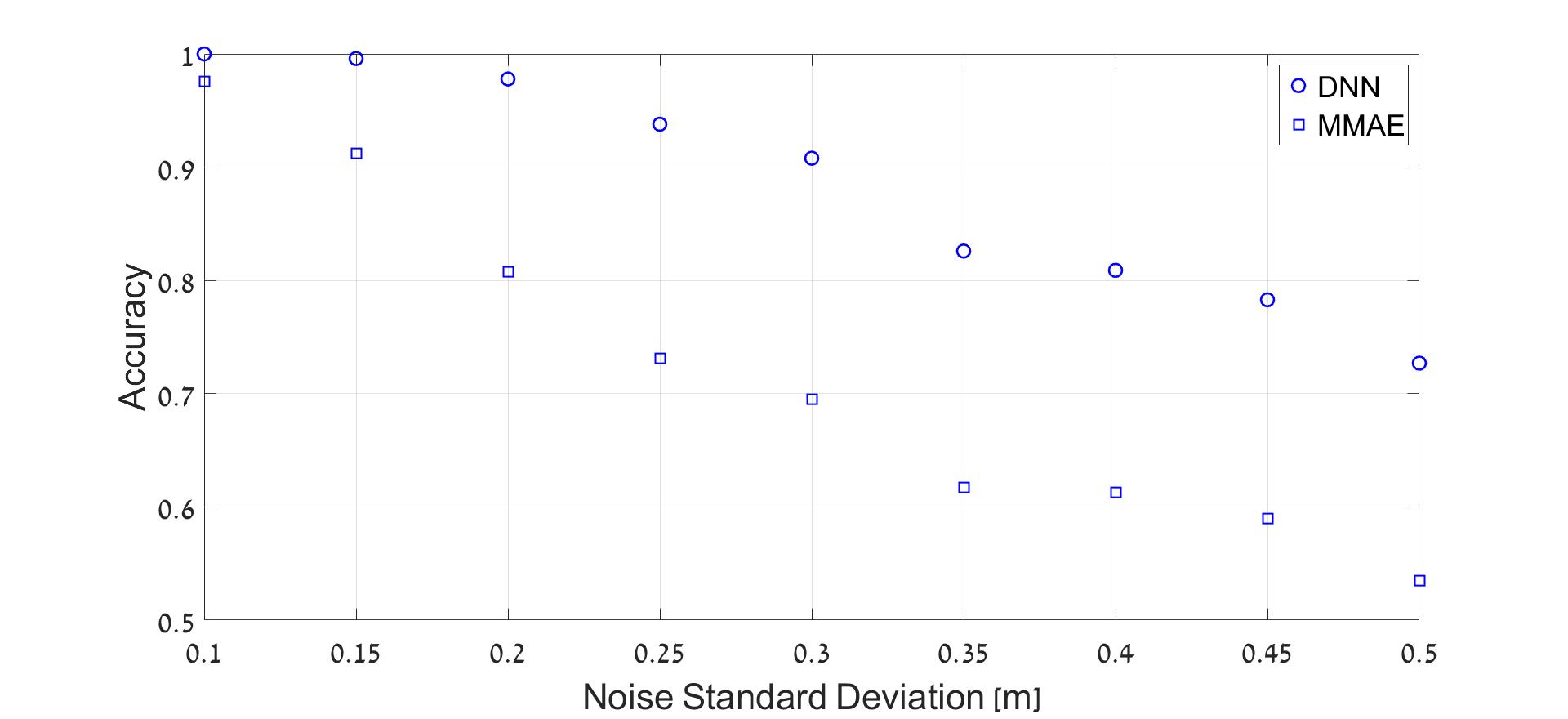} \\
  \caption{Accuracy vs noise standard deviation for \newline $\omega  \in \left\{ {10,10.2,10.4} \right\}\;\,[Hz]\;(\omega  \in {\Omega _3})$.}
  \label{fig:Acc_DNN_MMAE_10}
\end{figure}

 \figref{fig:Acc_DNN_MMAE_5} and  \figref{fig:Acc_DNN_MMAE_10} presents a comparison of the DNN and the MMAE estimators accuracy versus noise standard deviation, for $T=1 [s]$.
 For example, in \figref{fig:Acc_DNN_MMAE_5}, for a given noise standard deviation of $\sigma=0.3[m]$, the DNN classified correctly $99.5\% $ of the tested signals, while the MMAE classifier was successful only in $90.2\% $ of the test cases.

From \figref{fig:Acc_DNN_MMAE_5} and  \figref{fig:Acc_DNN_MMAE_10} it is evident that the DNN estimator consistently outperforms the MMAE estimator, for the examined set of frequencies.

\begin{figure}[!ht]
  \centering
  \includegraphics[width=9cm]{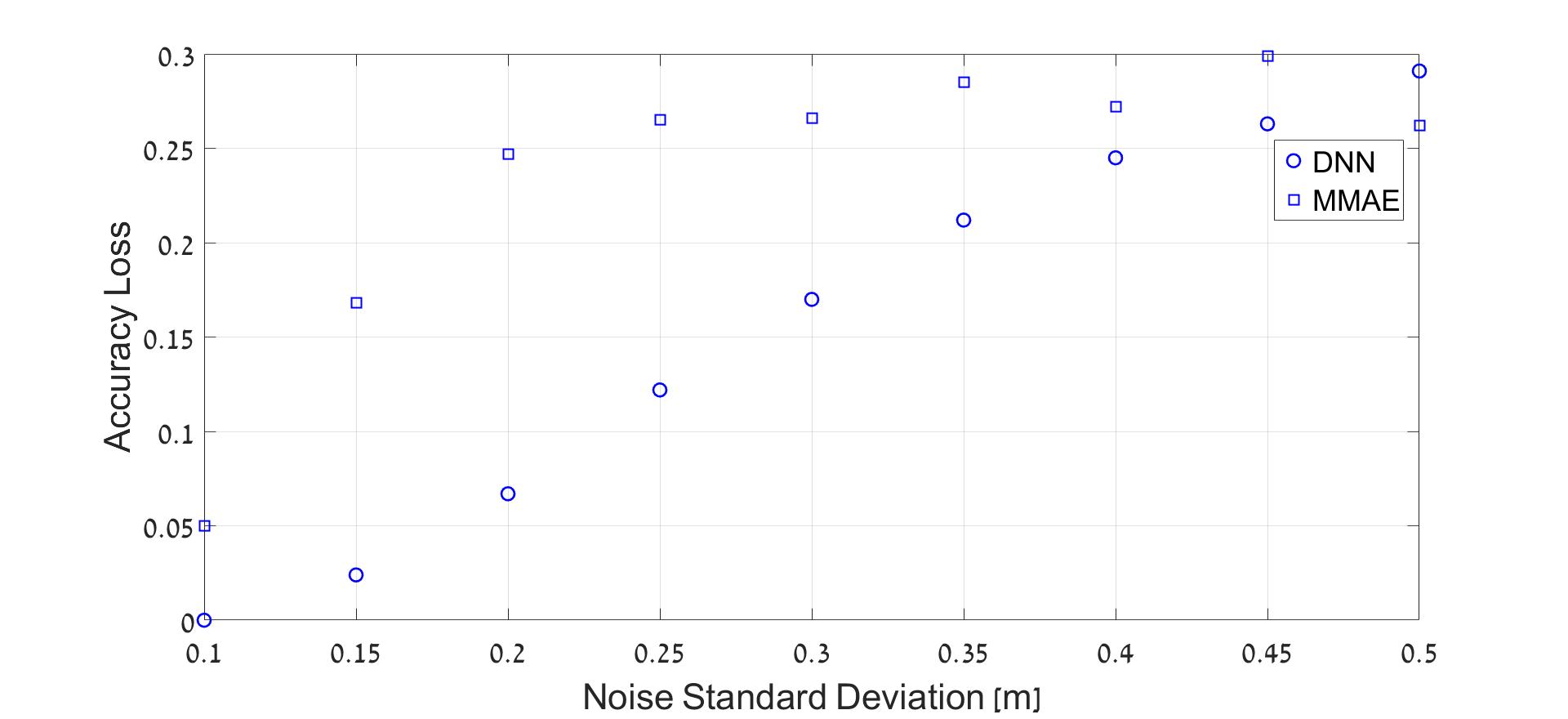} \\
  \caption{Accuracy loss vs noise standard deviation given a change from $\Omega_1$ to $\Omega_2$.}
  \label{fig:Acc_DNN_MMAE_5_5res}
\end{figure}

In order to analyze the sensitivity of the classification to the resolution of the frequencies in the given set $\Omega$, we compared the accuracy performance of set $\Omega_1$ (resolution of $0.5 [Hz]$) to the set $\Omega_2$ (resolution of $0.2 [Hz]$). The loss of accuracy, defined as the accuracy for set $\Omega_1$ minus the accuracy for set $\Omega_2$, was calculated and plotted versus the noise standard deviation. It is expected that for a given noise standard deviation, the accuracy will decrease (i.e. a positive accuracy loss) as the resolution is lowered.
\figref{fig:Acc_DNN_MMAE_5_5res} presents the results.

From  \figref{fig:Acc_DNN_MMAE_5_5res} it is evident that the performance of the DNN estimator is more robust to frequency resolution, i.e. it degrades gracefully, as the difference between the frequencies in the set becomes smaller.

\begin{figure}[!ht]
  \centering
  \includegraphics[width=9cm]{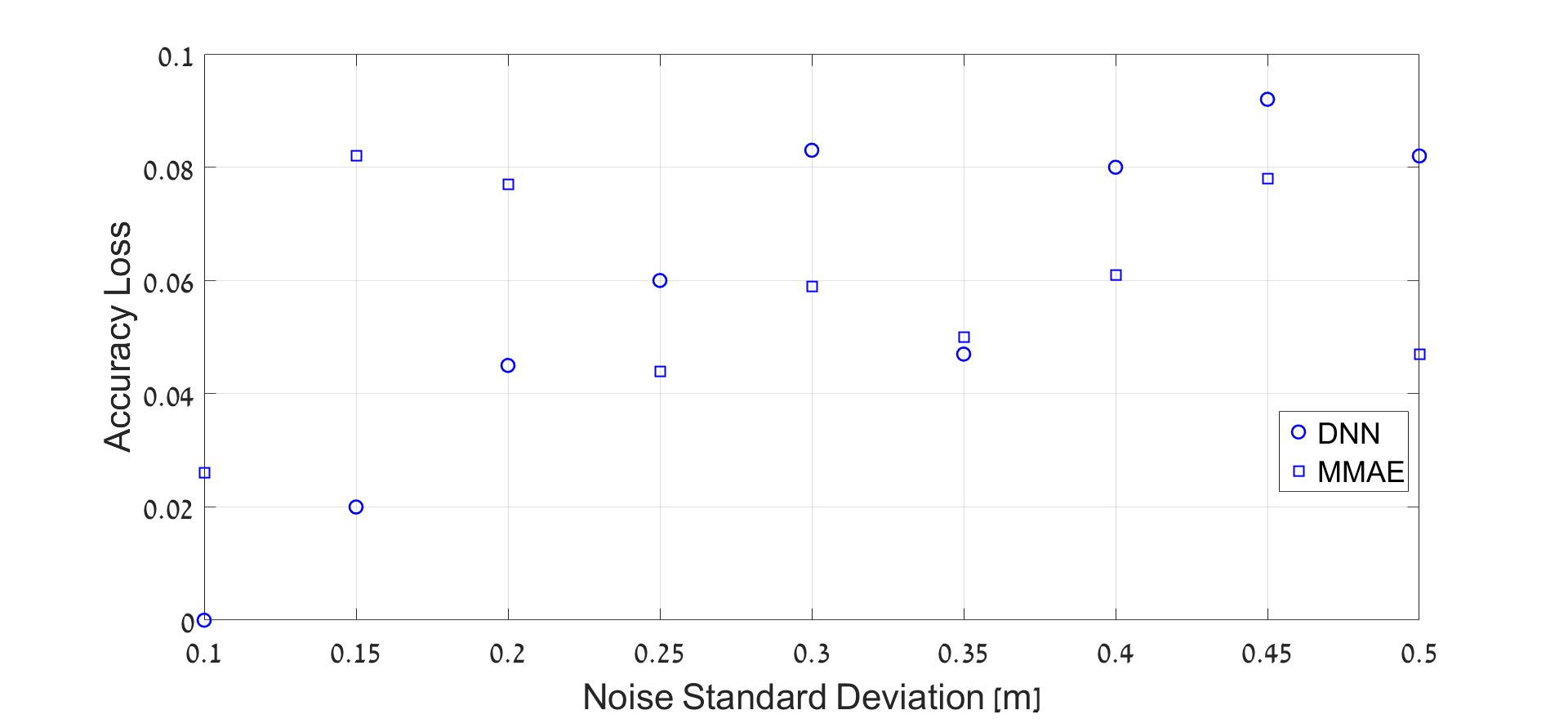} \\
  \caption{Accuracy loss vs noise standard deviation given a change from $\Omega_3$ to $\Omega_2$.}
  \label{fig:Acc_DNN_MMAE_10_5res}
\end{figure}

The sensitivity to the frequency value, given the same frequency difference in the set, is presented in \figref{fig:Acc_DNN_MMAE_10_5res}. The sensitivity to the frequency value was calculated by comparing the accuracy performance of set $\Omega_2$ to the performance of set $\Omega_3$, and the loss of accuracy, defined as the accuracy for set $\Omega_3$ minus the accuracy for set $\Omega_2$ was calculated and plotted versus the noise standard deviation.
\figref{fig:Acc_DNN_MMAE_10_5res} presents the results.

From  \figref{fig:Acc_DNN_MMAE_10_5res} it is evident that both the MMAE and the DNN classifiers perform better for higher values of frequencies in the frequency set, given the same difference between the frequencies in the set.

\begin{figure}[!ht]
  \centering
  \includegraphics[width=9cm]{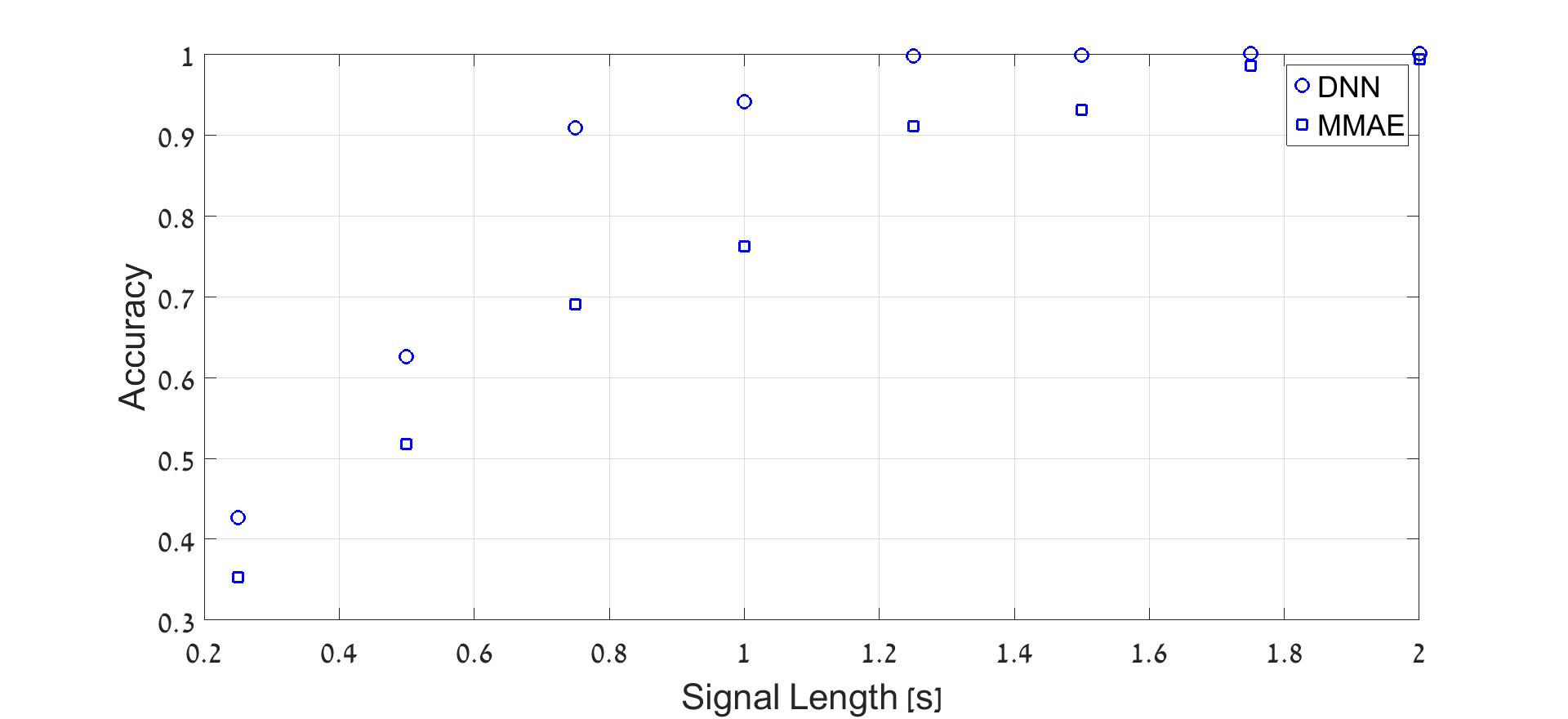} \\
  \caption{Accuracy vs signal length.}
  \label{fig:signal_STR}
\end{figure}

The length $T$ of the signal represents the number of measurements required for the estimation, as $T$ decreases the number of measurements also decreases, making the estimation process more challenging.
In order to evaluate the sensitivity of the results to signal length $T$, $T$ was changed, without changing the sampling time. \figref{fig:signal_STR} presents the results for  ${\Omega _1} = \left\{ {5,5.5,6} \right\}[Hz]$.

\figref{fig:signal_STR} suggest that the DNN estimator is able to achieve the same accuracy as the MMAE estimator, in a shorter signal length $T$. For example, to achieve an accuracy above $90\% $, the DNN requires a signal length of $0.75 [s]$, which is equivalent to $75$ measurements while the MMAE requires $1.25 [s]$, which is $125$ measurements, that is, more than $50\% $.

\begin{figure}[!ht]
  \centering
  \includegraphics[width=9cm]{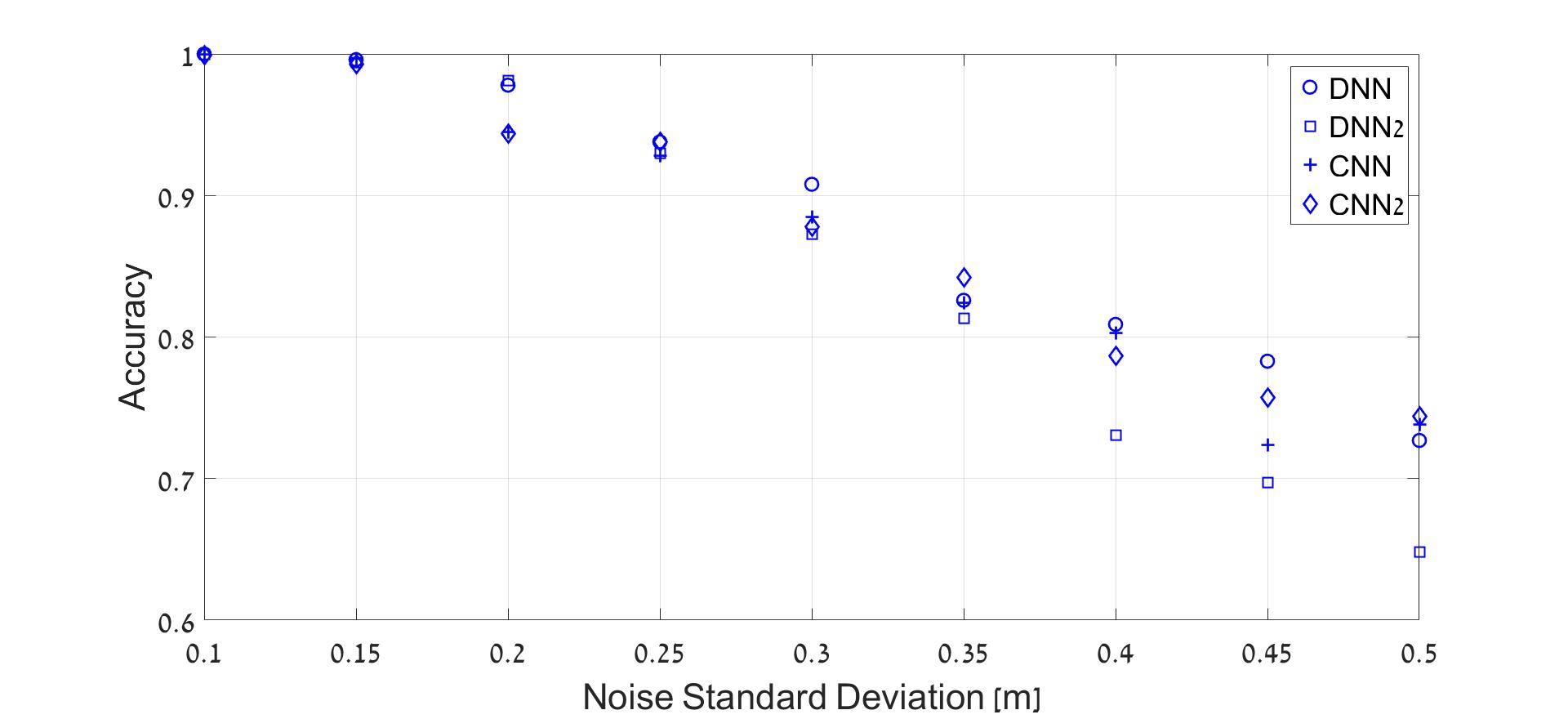} \\
  \caption{Accuracy vs noise standard deviation, for several NN architectures.}
  \label{fig:DNNandothers_acc}
\end{figure}

In order to compare the performance of several NN architectures, an accuracy comparison for different values of noise standard deviation was conducted for the set $\Omega_3$. The comparison is presented in \figref{fig:DNNandothers_acc}.
The compared architectures are described in \ref{subsec:Chosen DNN Architecture}

\figref{fig:DNNandothers_acc} indicates that the "DNN" architecture produces the highest overall performance, compared to other NN architectures. In addition, it is evident that overall, the variance of the accuracy between the architectures grows with the increase of noise standard deviation.


\section{Conclusions}
\label{sec:Conclusions}
In this paper we demonstrated the potential of using neural networks in target tracking applications. To that end, a DNN architecture was designed to estimate the unknown weaving target frequency instead of using a KF-based estimator.
We have demonstrated, using numerical simulations, that a DNN, applied to the classification of a maneuvering target frequency performs better compared to the MMAE estimator for the examined scenarios. The DNN predicted the target frequency with a higher accuracy than the MMAE estimator for a given noise standard deviation. Also, it was shown that for small time lengths of the signal (i.e. less measurements), DNN outperforms  MMAE. In addition, several fully connected and convolutional neural network architectures were designed and compared to each other in terms of accuracy performance, in which the DNN architecture labeled "DNN" obtained best overall performance.
In addition to a high accuracy, the DNN has a low computational load at the online stage due to the fact that the weights were determined at the offline training stage, thus making it an attractive alternative to the MMAE.

\bibliography{DL_sin}
\bibliographystyle{IEEEtran}

\end{document}